\documentclass{article}

\usepackage{times}  
\usepackage{helvet}  
\usepackage{courier}  
\usepackage{url}  
\usepackage{graphicx}  
\usepackage[margin=1in]{geometry}
\usepackage[utf8]{inputenc} 
\usepackage[T1]{fontenc}    
\usepackage{hyperref}       
\usepackage{url}            
\usepackage{booktabs}       
\usepackage{amsfonts}       
\usepackage{nicefrac}       
\usepackage{microtype}      

\usepackage{microtype}
\usepackage{booktabs}
\usepackage{wrapfig}
\usepackage{amssymb}
\usepackage{natbib}
\usepackage{algorithm}
\usepackage{algorithm, algorithmicx, algpseudocode}
\usepackage{balance}
\usepackage{subfig}%
\usepackage{amsmath}
\usepackage{verbatim}
\usepackage{commath}

\newcommand{\squishlist}{
 \begin{list}{$\bullet$}
  { \setlength{\itemsep}{0pt}
     \setlength{\parsep}{2pt}
     \setlength{\topsep}{2pt}
     \setlength{\partopsep}{0pt}
     \setlength{\leftmargin}{1em}
     \setlength{\labelwidth}{1em}
     \setlength{\labelsep}{0.5em} } }
     
     \newcommand{\squishend}{
  \end{list}  }
\DeclareFontFamily{U}{mathx}{\hyphenchar\font45}
\DeclareFontShape{U}{mathx}{m}{n}{
      <5> <6> <7> <8> <9> <10>
      <10.95> <12> <14.4> <17.28> <20.74> <24.88>
      mathx10
      }{}
\DeclareSymbolFont{mathx}{U}{mathx}{m}{n}
\DeclareFontSubstitution{U}{mathx}{m}{n}
\DeclareMathAccent{\widecheck}{0}{mathx}{"71}
\DeclareMathAccent{\wideparen}{0}{mathx}{"75}

\newtheorem{prop}{Proposition}

\title{Resource Constrained Deep Reinforcement Learning}
\author{Abhinav Bhatia, Pradeep Varakantham and Akshat Kumar}
\begin{document}

\maketitle

\begin{abstract}
In urban environments, supply resources have to be constantly matched to the "right" locations (where customer demand is present) so as to improve quality of life. For instance, ambulances have to be matched to base stations regularly so as to reduce response time for emergency incidents in EMS (Emergency Management Systems);  vehicles (cars, bikes, scooters etc.) have to be matched to docking stations so as to reduce lost demand in shared mobility systems.  Such problem domains are challenging owing to the demand uncertainty, combinatorial action spaces (due to allocation) and constraints on allocation of resources (e.g., total resources, minimum and maximum number of resources at locations and regions).

Existing systems typically employ myopic and greedy optimization approaches to optimize allocation of supply resources to locations. Such approaches typically are unable to handle surges or variances in demand patterns well. Recent research has demonstrated the ability of Deep RL methods in adapting well to highly uncertain environments. However, existing Deep RL methods are unable to handle combinatorial action spaces and constraints on allocation of resources. To that end, we have developed three approaches on top of the well known actor critic approach, DDPG (Deep Deterministic Policy Gradient) that are able to handle constraints on resource allocation. More importantly, we demonstrate that they are able to outperform leading approaches on simulators validated on semi-real and real data sets.
\end{abstract}

\section{Introduction}
This paper is motivated by aggregation systems that aggregate supply to improve efficiency of serving demand. Such systems have been employed in mobility systems, emergency response, logistics, food delivery,  grocery delivery, and many others.  There are multiple supply resources (e.g., ambulances, delivery/movement vehicles, taxis)  controlled by a central agency that need to be continuously allocated to supply entities (e.g., base stations, docking stations) so as to improve service efficiency for customer demand.  This sequential allocation problem becomes challenging due to combinatorial action space (allocating resources to entities), cost of reallocation, uncertainty in demand arrival,  constraints on resource allocation and in some cases also due to uncertainty in resource movement.

Existing systems typically employ myopic (single or few time steps) and greedy optimization approaches~\citep{Yue:2012,GhoshVAJ17,powell1996stochastic, lowalekar2017online} to optimize allocation of supply resources to locations.
As we demonstrate in our experimental results, greedy approaches  perform poorly when there are surges in demand or when variance in demand is high. Recent extension to employ Deep Learning with Reinforcement Learning, referred to as Deep RL, has significantly improved the scalability and effectiveness of RL in dealing with complex domains~\citep{mnih2015human,mnih2016asynchronous,Lillicrap15}. In this paper, we propose the use of Reinforcement Learning (RL) approaches to learn decisions in aggregation systems that can better represent and account for the sequential nature of decision making and uncertainty associated with demand.

However, current Deep RL methods are not directly suitable for handling aggregation systems of interest due to two reasons: (i) Deep RL methods do not scale well in domains with discrete and combinatorial action space, more so in problems at the scale of a city; (ii) Due to resource allocation constraints, action space is constrained. There have been research works that have provided mechanisms for solving resource allocation problems with Deep RL~\citep{Dulac-ArnoldESC15,MSR2016deepRM}. However, they do not consider constraints on resource allocation. ~\cite{AmosK17} have integrated quadratic optimization problems as individual layers in end-to-end trainable deep learning networks. Such networks (OptNet) could potentially be integrated with RL to handle resource constraints. Unfortunately, as indicated in their paper, they can only solve small problems due to the computational complexity of training these networks. ~\cite{Pham_2018} proposed an architecture (OptLayer) building on ideas from OptNet for constrained RL in context of robotics. They were able to demonstrate scaling to problems on a 6-DoF robot (6 dimensional action space). Unfortunately, their approach does not scale to problems of our interest where we have 95 dimensions.

DDPG (Deep Deterministic Policy Gradient)~\citep{Lillicrap15} is an approach that has been applied to multi dimensional continuous control problems with great results. However, like the other Deep RL methods, it is also unable to handle constraints on resource allocation. We propose extensions to DDPG that are able to handle constraints on resource allocation. We make five key contributions in this paper. First, we formally define the Resource Constrained Reinforcement Learning (ReCO-RL) model to represent problems of interest (decision making in aggregation systems). We are specially interested in hierarchical linear constraints, a useful subset of ReCO-RL problems. Second, we provide an extension to DDPG referred to as Constrained Projection (CP) that is generic (works for any kinds of resource constraints) but ensures constraints only approximately and has adhoc theoretical justifications. Third, we provide a novel, fast, scalable, simple to implement Constrained Softmax (CS) extension to DDPG that provably ensures constraints on resource allocation, but works only for a subset of hierarchical linear constraints. Next, we provide another novel, fast and scalable, Approximate Optlayer (ApprOpt) extension to DDPG that can provably handle any hierarchical linear constraints, while being orders of magnitude faster than OptLayer. Finally, we demonstrate that our extensions DDPG-CP, DDPG-CS and DDPG-ApprOpt provide either comparable or significantly better solutions than existing best approaches on two simulators for emergency response and bike sharing. Customer demand in these simulators was generated using real or semi-real datasets.

\section{Background}
In this section, we briefly describe the Reinforcement Learning (RL) problem~\citep{sutton1998reinforcement} and the Deep Deterministic Policy Gradient (DDPG) algorithm~\citep{silver2014deterministic} that is used to learn in environments with continuous action spaces. 

The RL problem  to maximize the long term reward while operating in an environment can be represented as a Markov Decision Process (MDP). Formally, an MDP is represented by the tuple $\big \langle \mathcal{S,A,T,R }\big \rangle$, where $\mathcal{S}$ is the set of states, $\mathcal {A}$ is the set of actions, $T(s,a,s')$ represents the stochasticity in the underlying environment and provides the probability of transitioning from state $s$ to state $s'$ on taking action $a$. $R(s,a)$ represents the reward obtained on taking action $a$ in state $s$. The RL problem is to learn a policy that maximizes the long term reward from \textit{experiences} without knowing the exact model of transitions and rewards. An experience is defined as a tuple $(s, a, s', r)$, and typically learning happens over a batch of experiences (referred to as an episode) that ends when $s'$ is a terminal state.  Q-learning  represents the value function for being in state $s$ and taking action $a$:
\begin{align}
&Q(s,a) = \mathbb{E}_{s',r} [r + \gamma \cdot \max_{a'} Q(s', a')] 
\end{align}
where the expectation, $\mathbb{E}$ is over the stochasticity in the environment with respect to transitions and also reward.

Since, we extend on Deep Deterministic Policy Gradient (DDPG) approach in this paper, we provide a brief description.  DDPG works in domains with continuous action spaces.  In DDPG, we have a \textit{critic} function parameterized by $\theta^Q$ that approximates the state-action-value function. We also have an \textit{actor} parameterized by $\theta^\mu$ that outputs the deterministic action in a continuous  space given the current state. Let $N$ denote the batch size or total experiences $e_i = (s_i, a_i, s_{i+1}, r_i)$ collected in an episode. The critic is updated by minimizing the loss:
\begin{align}
L &= \frac{1}{N} \sum_{i=1}^N (y_i - Q(s_i,a_i | \theta^Q))^2, \text{where} \nonumber\\
y_i &= r_i + \gamma \cdot Q'(s_{i+1}, \mu'(s_{i+1}|\theta^{\mu'})|\theta^{Q'}) \nonumber
\end{align}
where $Q'$, $\mu'$ are \textit{target} networks whose parameters lag behind the original networks $Q$, $\mu$. This is done to avoid making targets $y_i$ non-stationary, and improve the stability of updates. Next, actor policy $\mu$ is updated by using the sampled policy gradient:
\begin{align}
\nabla_{\theta^{\mu}} J &\approx \frac{1}{N} \sum_{i} \nabla_{a} Q(s, a | \theta^Q)|_{s = s_i, a = \mu(s_i)} \nabla_{\theta^{\mu}} \mu(s | \theta^{\mu})|_{s_i}  \nonumber
\end{align}
Target network parameters are updated periodically using the main network parameters.
\section{Resource Constrained Reinforcement Learning (ReCO-RL)}
\label{sec:recorl}

In our motivating problems of interest, supply resources have to be allocated online to the right entities to efficiently serve  demand. There is uncertainty in demand (w.r.t. both serving time and arrival) and potentially also in resource movement. Since allocation decisions at one stage have an impact on the subsequent stages and there is transitional uncertainty, RL is an ideal model for problems of interest in this paper.  However, a \textit{key differentiating factor} from typical RL problems is that the action space in problems of interest is constrained due on resource allocation constraints. 

We propose a modification to the RL model that can represent such constraints on resource allocation called \textit{Resource Constrained Reinforcement Learning} (ReCO-RL). To capture domains of interest, we have $n$ entities (e.g., base stations) where supply resources are situated and $m$ zones that capture customer demand.
Similar to the RL setting, the underlying tuple is $<S, A, T, R>$
\squishlist
\item
\noindent \textbf{States, S:}  Each state $s \!\in\! S$ is a tuple $\big< s_1, \ldots s_n, d_1, \ldots, d_m, t \big>$ where $s_k$ is the number of resources assigned to entity $k$, $d_z$ is the demand for resources in zone $z$, and $t$ is the decision epoch.
\item
\noindent \textbf{Actions, A:} Each action $a \!\in\! A$ is a tuple $\big<a_1, \ldots, a_k, \ldots, a_n \big>$ where 
$a_k$ represents the number of resources assigned to entity $k$. Depending on the domain, there can be different allocation constraints on the action, including but not limited to:
\begin{enumerate}
\item \textit{{Global sum constraint:}}  This enforces the global constraint on number of total resources available: 
$$ \sum_k {a_k} = C$$ 
\item \textit{{Local minimum and maximum bounds:}} These constraints enforce the minimum and maximum number of resources to be allocated to an entity. For a given entity $k$, $$\widecheck{C}_k \leq a_k \leq \widehat{C}_k$$
\begin{equation}\label{given_local_consts_cond}
    \forall k, \widecheck{C}_k, \widehat{C}_k \in [0, C]; \sum_{k} \widecheck{C}_k \leq C \leq \sum_{k} \widehat{C}_k 
\end{equation}
\item \textit{Regional minimum and maximum bounds:} These constraints enforce the minimum and maximum number of resources to be allocated to a region (a subset of entities). For a given region ${G}_j$, $\widecheck{C}_{{ G}_j} \leq \sum_{k \in { G}_j} a_k \leq \widehat{C}_{G_j}$ .
\begin{figure}[htbp]
\centering
\includegraphics[width=3in,height=1.5in]{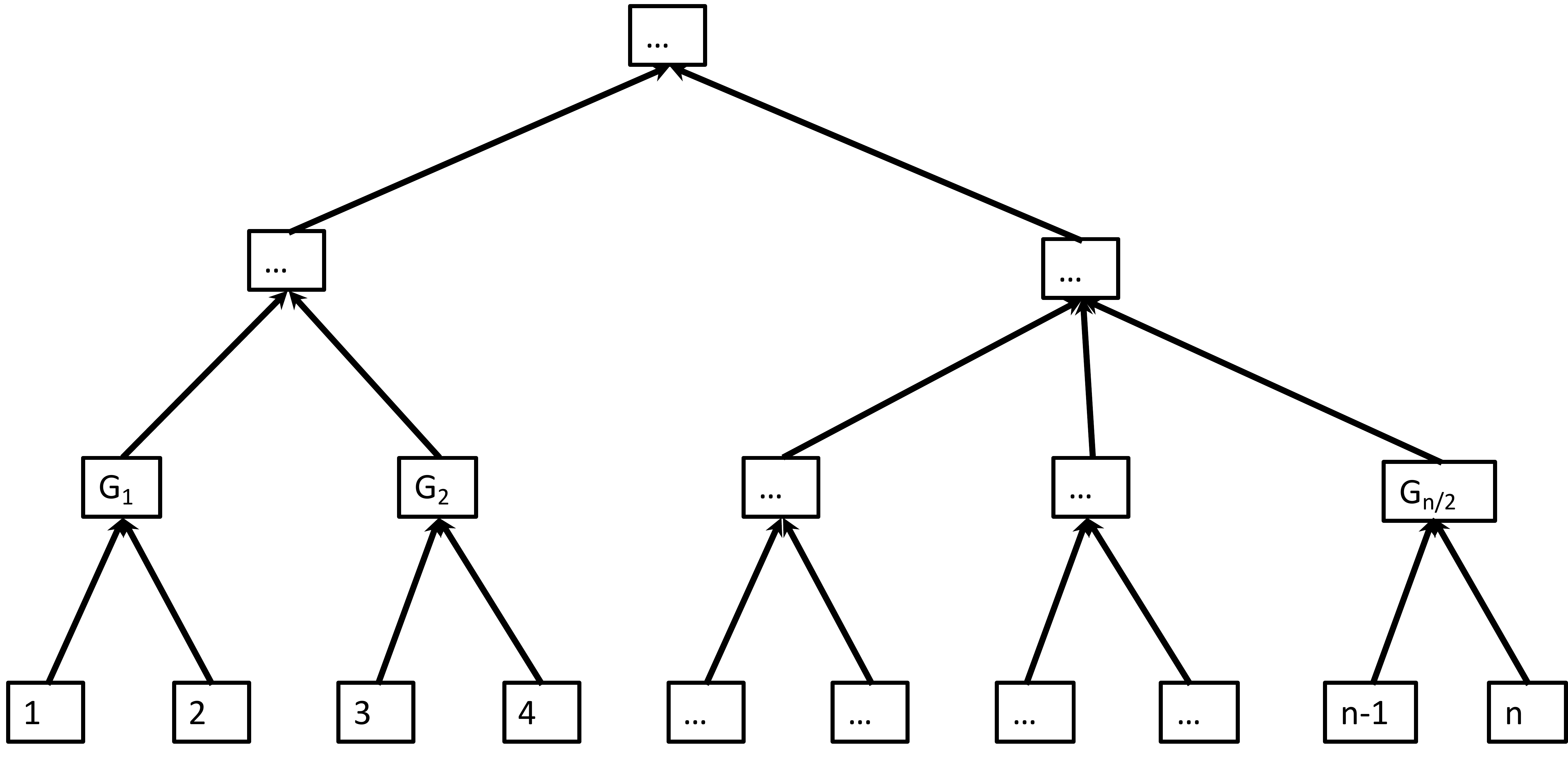}
\caption{\small Region Tree}
\label{fig:graph}
\end{figure}

In the most general case,  constraints can be on any subsets of regions. However, in practice and in problem domains of interest, there is a region hierarchy to ensure effective management.  For instance, in emergency response with ambulances:\\
\noindent (a) City is divided into multiple major regions (East, West, North, South, Centre);\\
\noindent (b) Each major region is divided into communities;\\
\noindent (c) Each community has some base stations;\\
\noindent The connection between regions and entities can in such cases be represented as a tree as shown in Figure~\ref{fig:graph}. Region $G_1$ consists of entities (base stations) 1 and 2. 
\end{enumerate}
\item
\noindent \textbf{Transitions and Rewards:} $T(s,a,s')$ captures uncertainty in demand and movement of resources between entities. $R(s,a,s')$ represents the demand served or the utility of serving the demand.
\squishend

\noindent We now provide two examples of how ReCO-RL can represent the problems of interest:
\noindent \textbf{Emergency Response as ReCO-RL:} Emergency Management Systems (EMSs) are tasked with reducing the response times for emergencies in many cities by using resources like ambulances, fire trucks etc. There are $n$ base stations (entities) where ambulances (or other resources) are placed and requests for ambulance can arise anywhere in the city that is divided into $m$ zones. The goal is to place the right number of ambulances at the base stations, so as to optimize bounded time response (number of requests served within bounded time)~\citep{Yue:2012}.  $s_{k}$ represents the number of ambulances at $k$th base station; $d_z$ represents the demand in $z$th zone. With respect to action, $a_k$ represents the number of ambulances to be assigned to $k$th base station. For bounded time response, reward is 1 for every request that is served within bounded time, zero otherwise. Transitions between states are dependent on demand patterns and action taken with respect to movement of ambulances.  

\noindent \textbf{Bike Placement as ReCO-RL:}  In bike placement problem, there are $n$ docking stations where bikes are placed and requests for bikes can arise at these docking stations (thus in this case $m=n$). The goal is to place the right number of bikes at the right docking stations at the right times, so as to reduce lost demand~\citep{GhoshVAJ17}. With respect to state, $s_{k}$ represents the number of bikes at $k$th docking station. $d_z$ represents the demand in $z$th zone. With respect to action, $a_k$ represents the number of bikes to be assigned to $k$th docking station. Reward is -1 for every lost customer due to lack of bikes at a docking station. Transitions between states are dependent on demand patterns and action taken with respect to movement of bikes.

\subsection{Extensions to ReCO-RL}
In the definition of ReCO-RL, we have considered a single type of resource and we do not distinguish between resources assigned to the same entity. However, it is easy to extend the model to consider multiple types of resources (e.g., multiple types of ambulances and bikes). We will have state features to be $s^{\tau}_k$ indicating the number of resources of type $\tau$ assigned to entity $k$. We will have a similar modification to action features, $a^{\tau}_k$ indicating the number of resources of type $\tau$ assigned to entity $k$. Constraints can then be defined on these new action features in a similar way.  As we show in Footnote 1, our approaches can still be applied, as state and actions can be converted to continuous space in a similar way. For purposes of easy explainability and since there are many domains which operate with single resource types, we focus on single resource type in this paper. 

\section{Approaches}

ReCO-RL problems have a discrete and combinatorial action space in problems of interest in this paper. For instance, even the simplest ambulance allocation problems considered in this paper have approximately $32^{25}$ possible actions. Due to the combinatorial action space and the presence of constraints on actions, existing approaches for Deep RL are not suitable.  Deep Deterministic Policy Gradient (DDPG) approach is also not directly applicable. However, in this paper, we propose novel extensions on top of DDPG to solve ReCO-RL problems effectively and efficiently.

For DDPG to be applicable for solving ReCO-RL problems, there are two key challenges:
\begin{enumerate}
\item Action space should be continuous and not discrete.
\item Address constraints on actions. Such constraints imply not every action obtained using actor network is feasible and furthermore, unconstrained exploration strategies (like Ornstein-Uhlenbeck process) are not applicable (as they result typically in violation of action constraints). 
\end{enumerate}

First, we consider the easier challenge of dealing with discrete action space. Action space is discrete and combinatorial in domains of interest due to the need for allocation of resources at every decision epoch. However, it is easy to approximate such discrete and combinatorial resource allocation actions into continuous actions. For instance, consider a discrete action $a = (10, 20, 30, 40)$ that represents 10 resources assigned to entity 1, 20 resources assigned to entity 2 and so on. This is (approximately) equivalent to $(0.1, 0.2, 0.3, 0.4)$, where $0.1$ refers to the fraction of resources assigned to entity 1, $0.2$ refers to the fraction of resources assigned to entity 2 and so on\footnote{In case of multiple resource types, we will normalize each resource type separately. For instance, if there are two resource types, then an action $(10, 20, 5, 10)$ -- which indicates 10 resources of type 1 to entity 1, 20 resources of type 1 to entity 1, 5 resources of type 2 to entity 1, 10 resources of type 2 to entity 2 -- gets converted to $(\frac{10}{30}, \frac{20}{30}, \frac{5}{15}, \frac{10}{15})$.}.  In this paper, we employ such a conversion. Since we convert to a continuous action space, the constraints also get normalized to be between 0 and 1. We refer to $\widecheck{\cal C}$ as the normalized lower bound of $\widecheck{C}$ (i.e. $\widecheck{\cal C} = \frac{\widecheck{C}}{C}$) and $\widehat{\cal C}$  (i.e. $\widehat{\cal C} = \frac{\widehat{C}}{C}$). In this this converted continuous action space, the action components thus must sum to ${\cal C} = \frac{C}{C} = 1$.

Addressing the second challenge of handling constraints on actions within DDPG is one of the key contributions of this paper.  We provide three methods in the context of DDPG:
\begin{enumerate}
\item Constrained Projection (CP):  For Reco-RL problems, the actor network of DDPG  generates infeasible actions. In this method, we employ penalties to train the actor network to generate feasible actions. In case the actor generates an infeasible action, then for the purpose of taking an action in the environment, we use the nearest projection of the infeasible action in the feasible action space, computed using a Quadratic Program (QP). For the purpose of training, the policy gradient is computed at the infeasible output of the actor network.
\item Constrained Softmax (CS): In this approach, we introduce modifications of the traditional softmax function as new layers in the actor network to ensure that it generates feasible actions. The layers are differentiable since they are essentially closed form expressions. These layers are part of the end to end backpropagation training of the actor. i.e. the policy gradient is computed at the output of these layers. These layers can handle a subset of local and regional constraints.
\item Approximate OptLayer (ApprOpt): In this approach, we introduce new differentiable layers based on ideas from OptLayer, but orders of magnitude faster. The speedup comes from solving the QP approximately in a semi closed-form semi-iterative fashion, which makes computing the gradients trivial. These layers are part of the end to end backpropagation training of the actor. These can handle the full breadth of local and regional constraints.
\end{enumerate}

The feasible action computed using the above methods is continuous. As in the Wolpertinger approach~\citep{Dulac-ArnoldESC15}, we round off the continuous action to the nearest discrete solution to act in the environment and to train the critic.

As indicated earlier, unconstrained exploration strategies (like Ornstein-Uhlenbeck process, which adds noise to the generated action) cannot be used in ReCO-RL problems because they can result in violation of action constraints. Therefore, we employ adaptive parameter noise~\citep{PlappertHDSCCAA17} for exploration in all the approaches. 

Now we will present each of the approaches in detail. For DDPG-CS and DDPG-ApprOpt, we will consider how to generate feasible actions in presence of local constraints only. We will later show in the supplementary how these local-constraints handling layers can be used to build a computation graph to handle regional constraints as well.

\subsection{Constrained Projection}

In this method, we employ penalties to train the actor network to generate feasible actions. Hence, it is not guaranteed to always generate a feasible action. In case the generated action is infeasible, then for the purpose of taking an action in the environment, we use the nearest projection of the infeasible action in the feasible action space, computed using a Quadratic Program (QP).

More specifically, the final layer in the actor network uses a $\tanh$ activation function that is scaled to give values from 0 to 1. $a_k = \mu_k(s_t | \theta^{\mu})$ refers to the output corresponding to resource allocation for entity $k$ from the actor network. The actor is trained to satisfy the allocation constraints by adding \textit{violation} cost penalty terms to the policy gradient equation. The violation cost $\nu$ is given by:
{\small \begin{align}
\nu(\vec a) &=  |1 - \sum_k^n{a_k}| + \sum_{G_j}\max{(0, \widecheck{\cal C}_{G_j} - \sum_{k \in G_j}{a_k} )} \nonumber\\
&+ \sum_{G_j}\max{(0, \sum_{k \in G_j}{a_k}-  \widehat{\cal C}_{G_j}  )}
\end{align}}
Due to this violation cost, the updated expression for sampled policy gradient is given by:
{\small \begin{align}
\bigtriangledown_{\theta^{\mu}} J &\approx \frac{1}{N} \sum_{i} \Big[\bigtriangledown_{a} Q(s, a | \theta^Q)|_{s = s_i, a = \mu(s_i)} \bigtriangledown_{\theta^{\mu}} \mu(s | \theta^{\mu})|_{s_i} \nonumber\\
& - \lambda \cdot \bigtriangledown_{\theta^{\mu}} \nu(a)|_{a=\mu(s_i)}\Big] \nonumber
\end{align}}
where $\lambda$ is tuned for each domain separately\footnote{We use $\lambda=10^3$ for emergency response domain and $\lambda=10^5$ for bike sharing domain, as those value yield the best performance.}. 

If $\vec a$ is infeasible, we identify the nearest $L_2$ projection, $\vec{z}$ that satisfies all the given constraints using the following QP:
{\small \begin{align}
    \textbf{Minimize} & \norm{\vec{z}-\vec{a}}_{L_2}, \textbf{subject to} \nonumber\\
    & \sum_k^n {z_k} = 1; \hspace{0.05in} \widecheck{\cal C}_{G_j} \leq \sum_{k \in G_j}{z_k} \leq \widehat{\cal{C}}_{G_j}, \forall G_j \nonumber
\end{align}}

Note that this method has very broad scope and can handle even overlapping (non-hierarchical) linear constraints. The main weakness of this approach is that the policy gradient is computed at the infeasible output of the actor. This can be a problem since the critic is trained on feasible actions, making the Q-value at infeasible actions theoretically undefined and practically ill defined. Thus this method is theoretically adhoc.

\subsection{Constrained Softmax}
In this method, we introduce new differentiable layers to the action network that are dependent on the type of constraints present in the problem. These new layers compute a softmax output over the actor network output while satisfying the allocation constraints, and hence we refer to these additional layers collectively as {\em constrained softmax} layers. These layers become part of the end to end backpropagation training of the actor. Since {\em constrained softmax} layers are dependent on the type of constraints, we describe the changes for each type of constraint separately. 

\subsubsection{Global Sum Constraint}
We will start with the simplest case of having just the global constraint, i.e., $ \sum_k {a_k} = 1$. This can be handled by having only one additional layer at the end of the actor network, i.e. the traditional softmax layer. This layer computes softmax over the actor network output to give the feasible action $\vec z$ as follows:
\begin{align}
z_k  &= \frac{y_k}{\sum_{k} y_k} \text{ where } y_k = e^{min(0, \mu_k(s_t | \theta^{\mu}))} \label{eqn:ydef}
\end{align}

\subsubsection{Local Minimum and Maximum Bounds} 
We now consider the local minimum and maximum bounds for each entity $k$. Like in the previous case, we just need one extra layer that is a modification of the softmax layer to handle local bounds. The goal here is to identify a function $z$ (with inputs $\vec y$ and bounds $\{\widecheck{\cal C}_k\}$ and $\{\widehat{\cal C}_k\}$) that satisfies the following properties:
\begin{align}
& \textit{ Min, max bounds: } \widecheck{\cal C}_k \leq z_k(\vec{y}, \{\widecheck{\cal C}_i\}, \{\widehat{\cal C}_i\} ) \leq \widehat{\cal C}_k \nonumber\\
& \textit{ Global sum constraint: } \sum_k z_k(\vec{y}, \{\widecheck{\cal C}_k\}, \{\widehat{\cal C}_k\} ) = 1 \nonumber\\
& \textit{ Monotonicity: }\frac{\partial z_k}{\partial y_k} \geq 0 \textbf{ and } \forall i \neq k, \frac{\partial z_k}{\partial y_i} \leq 0 \label{mon}
\end{align}
Monotonicity is required to identify the conditions where the functional form will yield a maximum. 
We have found one functional form that satisfies the properties above under some conditions. Proposition~\ref{eqn:prop} provides this functional form (which is a minor modification to the traditional softmax layer) along with the conditions under which they are applicable.

{ \begin{prop}
If we only have the sum constraint and local maximum bounds, $\{\widehat{\cal C}_k\}$ for individual entities, then the feasible outputs $\{z_k\}$ are given by:
{\small \begin{align}
 z_k(\vec{y}, \vec{0}, \{\widehat{\cal C}_k\} ) &= \frac{y_k + \epsilon_k}{\sum_{i} \Big[y_i + \epsilon_i\Big]} \text{  with } \epsilon_k =  \frac{\widehat{\cal C}_k \cdot (n-1)}{\sum_{i} \widehat{\cal C}_i -1} - 1 \label{locmaxeps} 
\end{align}}
where $\forall k, \epsilon_k \geq 0$, $ \sum_{i} \widehat{\cal C}_i  \neq 1$
\label{eqn:prop}
\end{prop}}

\noindent \textbf{Proof}: There are three steps to the proof:

\noindent \textit{\textbf{Step 1}}: \noindent $z_k$ satisfies monotonicity properties of \eqref{mon}:
{\small \begin{align}
\frac{\partial z_k}{\partial y_k} = \frac{ \sum_{i} \Big[y_i + \epsilon_i\Big] - (y_k + \epsilon_k)}{(\sum_{i} \Big[y_i + \epsilon_i\Big])^2} = \frac{\sum_{i \neq k} \Big[y_i + \epsilon_i\Big]}{(\sum_{i} \Big[y_i + \epsilon_i\Big])^2} \nonumber
\end{align}}
{\small \begin{align}
\frac{\partial z_k}{\partial y_i} = \frac{- (y_k + \epsilon_k)}{(\sum_{i} \Big[y_i + \epsilon_i\Big])^2} \nonumber
\end{align}}
Since $\forall i$, $y_i \geq 0$ (from Equation~\ref{eqn:ydef}) , $\frac{dz_k}{dy_k} \geq 0$ and $\frac{dz_k}{dy_i} \leq 0$ in all cases if we have $\forall i, \epsilon_i \geq 0$

\noindent \textit{\textbf{Step 2}}: Given the monotonicity properties of $z_k$, the maximum value for $z_k$ in Equation~\ref{locmaxeps} occurs when $y_k = 1$ and $\sum_{j \neq k} y_j = 0$. Therefore:
{\small \begin{align} 
&\frac{1 + \epsilon_k}{1 + \sum_{k} \epsilon_k} = \widehat{\cal C}_k \implies 1 + \epsilon_k = \widehat{\cal C}_k \cdot (1 + \sum_{k} \epsilon_{k}) \nonumber\\
\implies & \widehat{\cal C}_k \cdot \epsilon_1 + \ldots + (\widehat{\cal C}_k - 1) \cdot \epsilon_k + \ldots \widehat{\cal C}_k \cdot \epsilon_n = 1 - \widehat{\cal C}_k \label{eqn:axb}
\end{align}}
The above set of equations for all $k$ in matrix form is: ${\textbf{C}} \cdot \mathbf{\epsilon} =  \mathbb{C}$, which has a solution as long as determinant of ${\textbf{C}}$ is not zero. We calculate the determinant by performing the following steps: (i) subtract first column values from all the other columns; and (ii) add all rows to the first row. 
{ \begin{align}
|{\textbf{C}}|
= (\sum_{k} \widehat{\cal C}_k - 1) \cdot (\pm 1)
\label{det}
\end{align}}
Therefore, determinant is zero only if $\sum_{k} \widehat{\cal C}_k = 1$\footnote{When $\sum_{k} \widehat{\cal C}_k = 1$, then we assign $z_k = \widehat{\cal C}_k$, as that is the limit of $z_k$ when $\sum_{k} \widehat{\cal C}_k \to 1^{+}$. }

\noindent \textit{\textbf{Step 3}}:  Closed form expression for $\mathbf{\epsilon}$ can be verified by substituting in Equation~\ref{eqn:axb}. 
The $k^{th}$ row of $\textbf{C} \cdot \mathbf{\epsilon}$ is given by
{\small \begin{align}
& \widehat{\cal C}_k \cdot  \Big[\frac{\widehat{\cal C}_1 \cdot (n-1)}{\sum_{i} \widehat{\cal C}_i -1} - 1\Big]  + \ldots + (\widehat{\cal C}_k - 1) \cdot \Big[\frac{\widehat{\cal C}_k \cdot (n-1)}{\sum_{i} \widehat{\cal C}_i -1} - 1\Big] + \ldots  \nonumber\\
&= 1 - \widehat{\cal C}_k
\end{align}}
This is the $k^{th}$ row of $\mathbb{C}$ and hence the expression. $\hfill \blacksquare$

For local minimum bounds, we allocate each entity its minimum bounds and solve the problem for the remaining value (1 - sum of minimum bounds) by normalizing the bounds. For instance, in a problem with 3 entities with minimum bounds as (0.1, 0.1, 0.1) and maximum bounds as (0.4, 0.5, 0.6). We convert it to a problem with minimum bounds as (0, 0, 0) and maximum bounds as (0.3/0.7, 0.4/0.7, 0.5/0.7). When there are both local minimum and maximum bounds, the expression for $z_k$ is: 
{\small \begin{align}
 z_{k}(\vec{y}, \{\widecheck{\cal C}_i\},\{\widehat{\cal C}_i\}) = \widecheck{\cal C}_k + (1 - \sum_{j} \widecheck{\cal C}_j) z_{k}(\vec{y}, \vec{0},\{\frac{\widehat{\cal C}_i - \widecheck{\cal C}_i}{1 - \sum_{j} \widecheck{\cal C}_j} \})
 \label{eqn:zk}
\end{align}}
Therefore, we can use Proposition~\ref{eqn:prop} to compute the expression for minimum bounds as well. 

In general, if the sum constraint is $\cal{C}$, then the expression for $z_k$ is:
{\small \begin{align}
    z_{k}(\vec{y}, \{\widecheck{\cal C}_i\},\{\widehat{\cal C}_i\}, {\cal C}) = \widecheck{\cal C}_k + ({\cal C} - \sum_{j} \widecheck{\cal C}_j) z_{k}(\vec{y}, \vec{0},\{\frac{\widehat{\cal C}_i - \widecheck{\cal C}_i}{{\cal C} - \sum_{j} \widecheck{\cal C}_j} \})
    \label{eqn:zk_general}
\end{align}}

\subsubsection{Limitation of Constrained Softmax}
The proof of proposition \eqref{eqn:prop} requires that $\forall k: \epsilon_k \ge 0$. i.e.
\begin{align}
    \forall k: \frac{\widehat{\cal C}_k \cdot (n-1)}{\sum_{i} \widehat{\cal C}_i -1} - 1 \ge 0 \label{eq:csmax}
\end{align}
An example when this condition will be violated is when $\{\widehat{\cal C}\}=(0.3,0.5,0.6)$. Therefore, the applicability of the constrained softmax is limited to cases where~\eqref{eq:csmax} is satisfied.

\subsection{OptLayer}

One way to address different constraints on the actions is to project the output of the actor neural network to the feasible space of actions. Assume that  $y_i\; \forall i$ is the output of the actor network, which may not satisfy all the constraints in our domain. We can project $\vec{y}$ to the feasible space and get the feasible action $\vec{z}$ by solving the following QP:
\begin{equation} \label{qp}
\small
    \begin{aligned}
        \underset{\vec{z}}{\min} \sum_{k=1}^n (z_k - y_k)^2 \quad &\text{subject to} \\[-10pt]
        \sum_{k=1}^n z_k - {\cal C} &= 0                   \qquad: \lambda\\[-5pt]
        \forall k=1..n : z_k - \widehat{{\cal C}}_k &\le 0      \qquad: \alpha_i  \\
        \forall k=1..n : \widecheck{{\cal C}}_k - z_k &\le 0    \qquad: \beta_i
    \end{aligned}
\end{equation}
Here $\lambda,\alpha_i,\beta_i$ are the corresponding Lagrange multipliers. Since the objective function is \textit{strictly} convex and the constraints are linear, there exists a unique solution to this QP.
The Lagrangian function is~\citep{Bertsekas99}:
{\small
\begin{align}
    L(\vec{z},\vec{\alpha},\vec{\beta},\lambda) &= \sum_k (z_k-y_k)^2 + \lambda (\sum_k z_k - {\cal C}) \nonumber\\
    &+ \sum_k \alpha_k(z_k - \widehat{{\cal C}}_k) + \sum_k \beta_k(\widecheck{{\cal C}}_k - z_k)
\end{align}}
The KKT conditions (conditions satisfied by the optimal solution $\vec{z}^\star, \vec{\alpha}^\star, \vec{\beta}^\star, \lambda^\star$) of the QP~\eqref{qp} are given by
{\small
\begin{align*}
    &\nabla_{\vec{z},\lambda} L & = \vec{0} \\
    \forall k=1..n : \quad &\alpha_k^\star(z_k^\star - \widehat{{\cal C}}_k) & = 0 \\
    \forall k=1..n : \quad &\beta_k^\star(\widecheck{{\cal C}}_k - z_k^\star) & = 0
\end{align*}}
In the backward pass, we need to compute the gradients of the optimal solution of the QP w.r.t.$\!$ the inputs $\vec{y}$ or  $J_{kj}\!=\! \partial z_k^\star/ \partial y_j $. Such gradients can be computed using the implicit function theorem as shown in~\citep{AmosK17}. Briefly, differentiating each of the above equations with w.r.t each input $y_j$ yields a system of system of linear equations in partials $\partial z_k^\star/ \partial y_j$, which can be solved to find the Jacobian matrix $J$. Thus in this method, the forward pass requires solving a QP using an optimizer and the backward pass involves solving the set of linear equations for each item in the sampled minibatch.

\subsection{Approximate OptLayer algorithm}

The main challenge in solving~\eqref{qp} in the forward pass is that it becomes computationally slow given the large number of iterations most RL approaches require. Therefore, we next propose an approximate algorithm to solving the QP~\eqref{qp}, which apart from being very efficient, makes computing the gradients in the backward pass computationally much faster. The motivation for our proposed approach comes from similar iterative approaches that have been used to solve QPs, but in different contexts such as graphical models~\citep{KumarZ11,DuchiSSC08}.

Before we describe the algorithm, we impose the condition that the output $\vec{y}$ of the neural network always satisfies respective upper/lower bounds or  $\widecheck{\cal C}_k \leq y_k \leq \widehat{\cal C}_k \; \forall k$. This condition is  easy to enforce in a neural network. Assume $\vec{x}=\mu_k(s_t | \theta^{\mu})$ is any arbitrary vector output of the network, we set $y_k \!=\! \widecheck{\cal C}_k + (\widehat{\cal C}_k - \widecheck{\cal C}_k) \frac{x_k-\min(\vec{x})}{\max(\vec{x}) - \min(\vec{x})}$, where $\min$, $\max$ provide minimum, maximum component of the vector respectively. We do this scaling only if any  $x_k$ violates its corresponding bounds. Else, we set $y_k=x_k$.

The main motivation behind this approach is that we do not need the optimal L2 projection of the input $\vec{y}$. We just need \textit{any differentiable projection} function which can respect the constraints. The term "approximate" is used because the projection derived by our proposed approach is not very far from the nearest L2 projection (as empirically observed). 
The algorithm proceeds by solving the QP assuming that the inequality constraints (or the lower and upper bound constraints in QP~\eqref{qp}) will be inactive . It can be shown that the solution to the QP after ignoring the inequalities is:
\begin{equation}\label{qp_closed_form}
    \forall k: z_k = y_k + ({\cal C} - \sum_{k=1}^n y_k) / n
\end{equation}
After applying this formula, if all the $z_k$ are found to be satisfying the constraints $\widecheck{\cal C}_k \le z_k \le \widehat{\cal C}_k$, then the algorithm terminates and the gradients are easily computed as $\partial{z_k}/\partial{y_j}=\delta_{kj} - 1/n$, where $\delta_{kj}$ is 1 if $k=j$, else 0. 

If some bound constraints are not satisfied for some outputs, the algorithm proceeds to correct the output $z$ in two phases: LOWER: to satisfy min constraints and UPPER: to satisfy max constraints. In the LOWER phase, all the outputs $z_k$ which are \textit{below} the respective min bounds are set equal to (or clamped to) the respective min bound $\widecheck{\cal C}_k$. The remaining value ${\cal C'}$ ($={\cal C}$ minus sum of clamped outputs) is redistributed over the remaining $n'$ ($=n$ minus number of clamped outputs) outputs using an expression similar to ~\eqref{qp_closed_form}, but with new ${\cal C}={\cal C'},n=n'$ and involving only the $y_k$ corresponding to the unclamped $z_k$. The LOWER phase loops until there is no need for anymore clamping i.e. it ends when $\forall k: z_k \ge \widecheck{\cal C}_k$.

After the LOWER phase is over, the UPPER phase begins, which similarly repeatedly clamps the outputs which violate the respective max constraints and redistributes the remaining value. During the entire process, if a $z_k$ is clamped either to its upper or lower bound, it remains clamped to the same value for all the future iterations and remains out of the redistribution equation along with its corresponding $y_k$. The derivative of such a clamped $z_k$ is $0$ w.r.t all inputs $y_j$, and derivative of all outputs $z_j$ is $0$ w.r.t such inputs $y_k$. Algorithm~\ref{algo_apprx_qp} provides the pseudocode.

\begin{algorithm}[t]
    \small
        \caption{Forward pass and gradient computation for ApprOpt layer}\label{algo_apprx_qp}
        \begin{algorithmic}[1]
            \Require $n \ge 2$
            \Require $\forall k=1..n: 0 \le \widecheck{\cal C}_k < \widehat{\cal C}_k \le {\cal C}$
            \Require $\sum \widecheck{\cal C}_k < {\cal C} < \sum \widehat{\cal C}_k$
            \Require $\forall k: \widecheck{\cal C}_k \le y_k \le \widehat{\cal C}_k$
            \State $n' \gets n$ \Comment{count of unclamped indices}
            \State $\cal C' \gets \cal C$ \Comment{value to distribute to unclamped indices}
            \State $\Omega \gets \{1,2,...,n\}$ \Comment{unclamped output indices}
            \State $\text{phase} \gets \text{LOWER}$ \Comment{possible values are: $\text{LOWER} = 0; \text{UPPER}=1; \text{DONE}=2$}
            \While{$\text{phase} \ne \text{DONE}$}
                \State $\Omega' \gets \phi$ \Comment{indices clamped in this iteration of the while loop}
                \For{$k \in \Omega$}
                    \State $z_k \gets y_k + ({\cal C'}-\sum_{j \in \Omega} y_j)/n'$
                    \State $J_{kj} \gets \delta_{kj} - 1/n' \ \mathbf{foreach}\  j \in \Omega$
                    \State $J_{kj} \gets 0 \ \mathbf{foreach}\  j \notin \Omega$
                    \If{$z_k < \widecheck{\cal C}_k\  \textbf{and}\  \text{phase}=\text{LOWER}$}
                        \State $z_k \gets \widecheck{\cal C}_k$
                        \State $J_{kj} \gets 0 \ \mathbf{foreach}\  j=1..n$
                        \State $\Omega' \gets \Omega' \cup \{k\}$
                    \ElsIf{$z_k > \widehat{\cal C}_k \ \textbf{and} \  \text{phase}=\text{UPPER}$}
                        \State $z_k \gets \widehat{\cal C}_k$
                        \State $J_{kj} \gets 0 \ \mathbf{foreach}\  j=1..n$
                        \State $\Omega' \gets \Omega' \cup \{k\}$
                    \EndIf
                \EndFor
    
                \State $n' \gets n'- |\Omega'|$
                \State ${\cal C'} \gets {\cal C'} - \sum_{k \in \Omega'} z_k$
                \State $\Omega \gets \Omega - \Omega'$
                \If{$\Omega' = \phi$}
                    \State $\text{phase} = \text{phase} + 1$
                \EndIf
            \EndWhile
            \State \Return $z, J$
            \Ensure $\forall k: \widecheck{\cal C}_k \le z_k \le \widehat{\cal C}_k$
            \Ensure $\sum_{k=1}^n z_k = \cal C$
            \Ensure $z = y \ \mathbf{if}\ \sum_{k=1}^n y_k = \cal C$
        \end{algorithmic}
\end{algorithm}

\vskip 2pt
\noindent{\textbf{Proof Sketch for Algorithm 1: }There are five steps to proving that the algorithm terminates and always gives a feasible output. All steps are proved in the supplementary in detail. \textit{\textbf{Step 1:}} Line 8 of algorithm \ref{algo_apprx_qp} always leaves at least one unclamped output in LOWER phase. \textit{\textbf{Step 2:}} LOWER phase always terminates, with not having already excluded all feasible solutions. \textit{\textbf{Step 3:}} Line 8 of algorithm \ref{algo_apprx_qp} always leaves at least one unclamped output in UPPER phase. \textit{\textbf{Step 4:}} UPPER phase always terminates. \textit{\textbf{Step 5:}} By step 4, the while loop terminates with $n'\ge 1$ and with a feasible output.

Whenever the input is feasible, i.e. $\sum_k y_k={\cal C}$ and $\widecheck{\cal C}_k \le y_k \le \widehat{\cal C}_k$, then $z_k=y_k$ by equation~\eqref{qp_closed_form} i.e. the algorithm behaves as an identity function for feasible inputs. Thus there exist a set of inputs such that the corresponding set of outputs occupies the entire feasible solution space.

\vskip 2pt
\noindent{\textbf{Computing gradients: }}Notice that algorithm~\ref{algo_apprx_qp} computes all the required gradients $J_{kj} \!=\! \partial z_k/\partial y_j$ in lines 9, 10, 13, and 17. There is no expensive matrix inversion or solving of a system of linear equation required. This provides significant speedup over the exact solving of QP using CPLEX and then computing the gradients as in the standard OptLayer.

\vskip 2pt
\noindent{\textbf{Practical Considerations:}} ~\cite{Pham_2018} found that using only the OptLayer gradient does not lead to efficient learning, since the actor network output is very far from the feasible actions in the beginning. We also found the same problem and hence we penalize the original actor network output $\vec x$ for being infeasible. But instead of using reward shaping as in~\cite{Pham_2018}, we include a penalty term in the actor's training objective, like in DDPG-CP. This penalization method gives direct gradient information to the actor network, as opposed to the critic learning the penalty returns first and then passing the gradient information back to the actor network. Even though we use a penalty in the actor's objective, this approach is still superior to DDPG-CP, since it is an end to end learning approach, and the policy gradients are propagated via the ApprOpt layer, making it theoretically more justified.

\section{Experiments}

Our goal in the experiments is to evaluate the performance of our new approaches, DDPG-CP, DDPG-CS and DDPG-ApprOpt in comparison to baseline approaches\footnote{To ensure reproducibility of the results, we provide a very detailed experimental section in the supplementary document. This includes specific details of the simulators, constraints, deep learning hyperparameters and other relevant details which are not provided in this main document due to space constraints}. DDPG-OptLayer was at least an order of magnitude slower and on the limited computation resources available to us, we were unable to evaluate it completely. Within the limited evaluations, we observed that DDPG-ApprOpt was on par with DDPG-OptLayer.

We train and evaluate our approaches on two simulators\footnote{\noindent The OpenAI Gym environments for the two simulators are available at: Emergency Response: \href{https://github.com/bhatiaabhinav/gym-ERSLE}{\textit{https://github.com/bhatiaabhinav/gym-ERSLE}}  Bike Sharing: \href{https://github.com/bhatiaabhinav/gym-BSS}{\textit{https://github.com/bhatiaabhinav/gym-BSS}}} related to emergency response and bike sharing that have inherent constraints. 

\noindent \textbf{Emergency Response:} First, we consider a simulator for emergency response with the transitional dynamics inspired by ~\cite{Yue:2012}. In our simulator, there are 32 ambulances to be distributed among 25 base stations. The RL agent is expected to act by providing a target-allocation every 30 minutes.  

For our experiments, we consider 2 demand patterns: "Singapore Poisson" (in which the probability of an incident at a place is a function of the time of the day, inspired by actual demand statistics in Singapore 2011~\citep{sgPattern2011}), and a "random surges" version in which we introduce unpredictable random surges in demand at random places at random times, once per episode. 

\noindent \textbf{Bike Sharing:}
We consider the simulator described in ~\cite{GhoshICAPS17} for Hubway bike sharing system. The bike sharing system consists of 95 base stations (zones) and 760 bikes.
The RL agent is expected to act by providing a target allocation every 30 minutes. The simulator may not be able to achieve this allocation at the end of the next 30 minutes, since in this domain, the allocation is influenced also by the customers picking up bikes at one zone and leaving them at different zones.

For our experiments we consider the first demand data set from ~\cite{GhoshICAPS17}, which corresponds to a slightly modified real demand data for 60 weekdays.
Like in that work, the dataset is divided as 20 days for training and 40 days for testing. Corresponding to these, we create a training MDP and a testing MDP, which use training data and testing data respectively. There are non-uniform local constraints in this environment.

\noindent \textbf{Baselines: } For ERS domain, we compare our approaches against static allocation baselines, computed offline using a greedy allocation algorithm, described in ~\cite{Yue:2012}. Greedy algorithm cannot consider constraints while calculating the allocation and hence the results from baselines appear stronger. It should be noted that since we consider allocation problems, we cannot employ planning methods like the one by ~\cite{fern}, where the focus is on planning the dispatch rather than allocation.

For bike sharing domain, we take the best results from ~\cite{GhoshICAPS17}, which correspond to "RTrailer": a dynamic repositioning algorithm using 10 bike trailers, each having a capacity of 5.

\noindent \textbf{Algorithmic Details: } For the actor and critic deep learning networks, we largely retain the basic architecture used in ~\cite{Lillicrap15}. Both the actor and the critic use two hidden layers with 128, 96 ReLU units for emergency response domain and 400,300 ReLU units for bike sharing domain. For the critic, actions are not included until the second hidden layer. Layer normalization is applied before non-linearities to all hidden layers in both the actor and the critic. Adaptive parameter noise~\citep{PlappertHDSCCAA17} is used for exploration with a target divergence of $1/C$, and adaptation factor of $1.05$. More detailed procedures and hyperparameters are provided in the supplementary section. 



For DDPG-CP, we use $\lambda=10^3$ for emergency response domain and $\lambda=10^5$ for bike sharing domain.
For DDPG-ApprOpt, we use penalty term coefficient of $10^3$ for Emergency Response Domain and $10^4$ for Bike Sharing.


\noindent \textbf{Training and Evaluation: } Each experiment consisted of training on the corresponding environment 5 times for 10,000 episodes using random seeds=0..4 to initialize the environment and model parameters. During the training, every 4th episode was played without exploration. These exploitative episodes were used to generate the learning curves. Each learning curve shown in figure \ref{fig:learning-curves} shows the mean and standard deviation of the smoothed individual learning curves across the random seeds. For evaluating a trained model, its average score was taken across 100 test episodes without exploration (simulator initialized with the random seed = 42). The final evaluation score for an experiment was calculated as the average score of the 5 trained models corresponding to it.

\subsection{Results and Discussion}

\begin{figure*}[htbp]
    \includegraphics[scale=0.19]{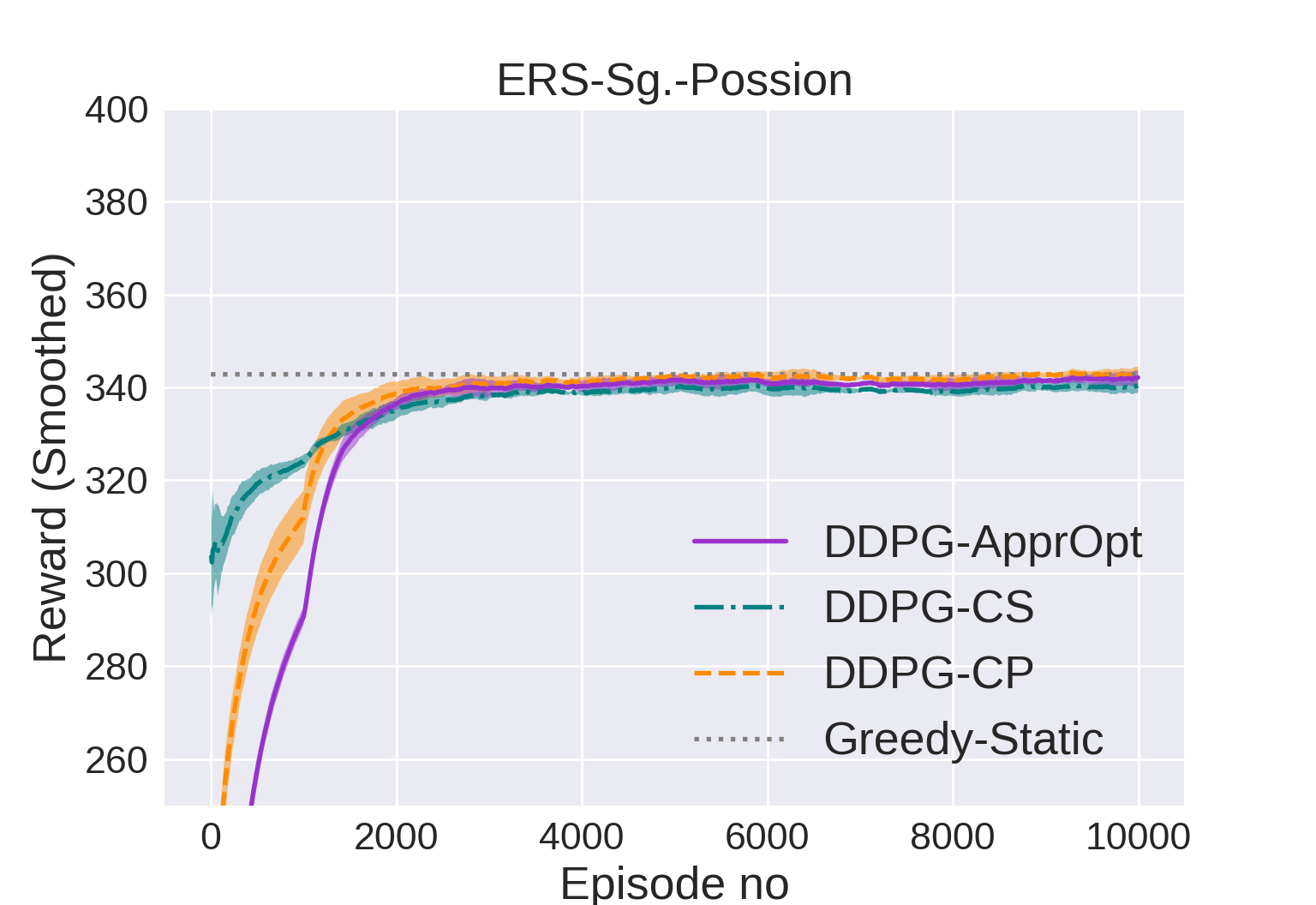}
    \includegraphics[scale=0.19]{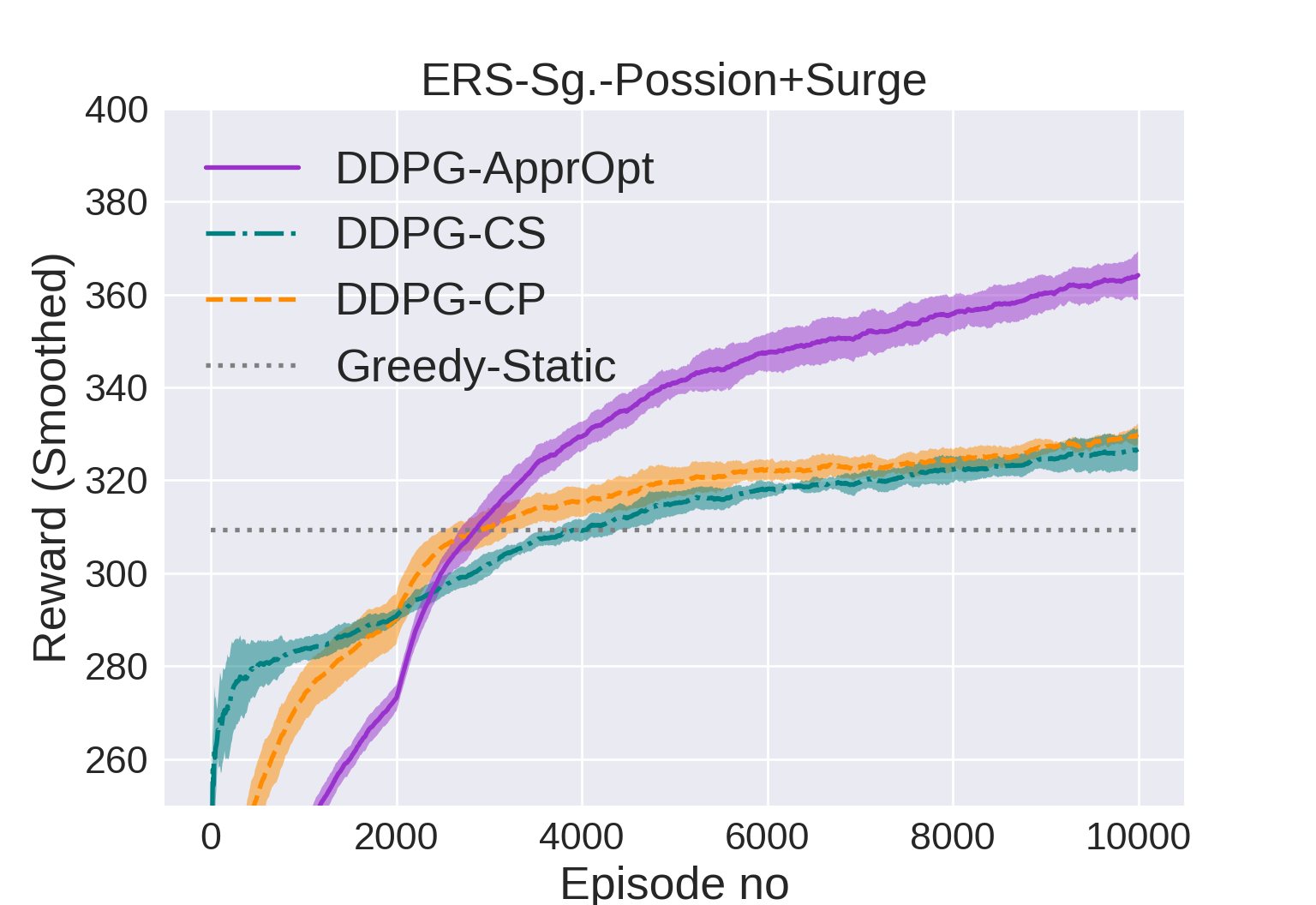}
    \includegraphics[scale=0.19]{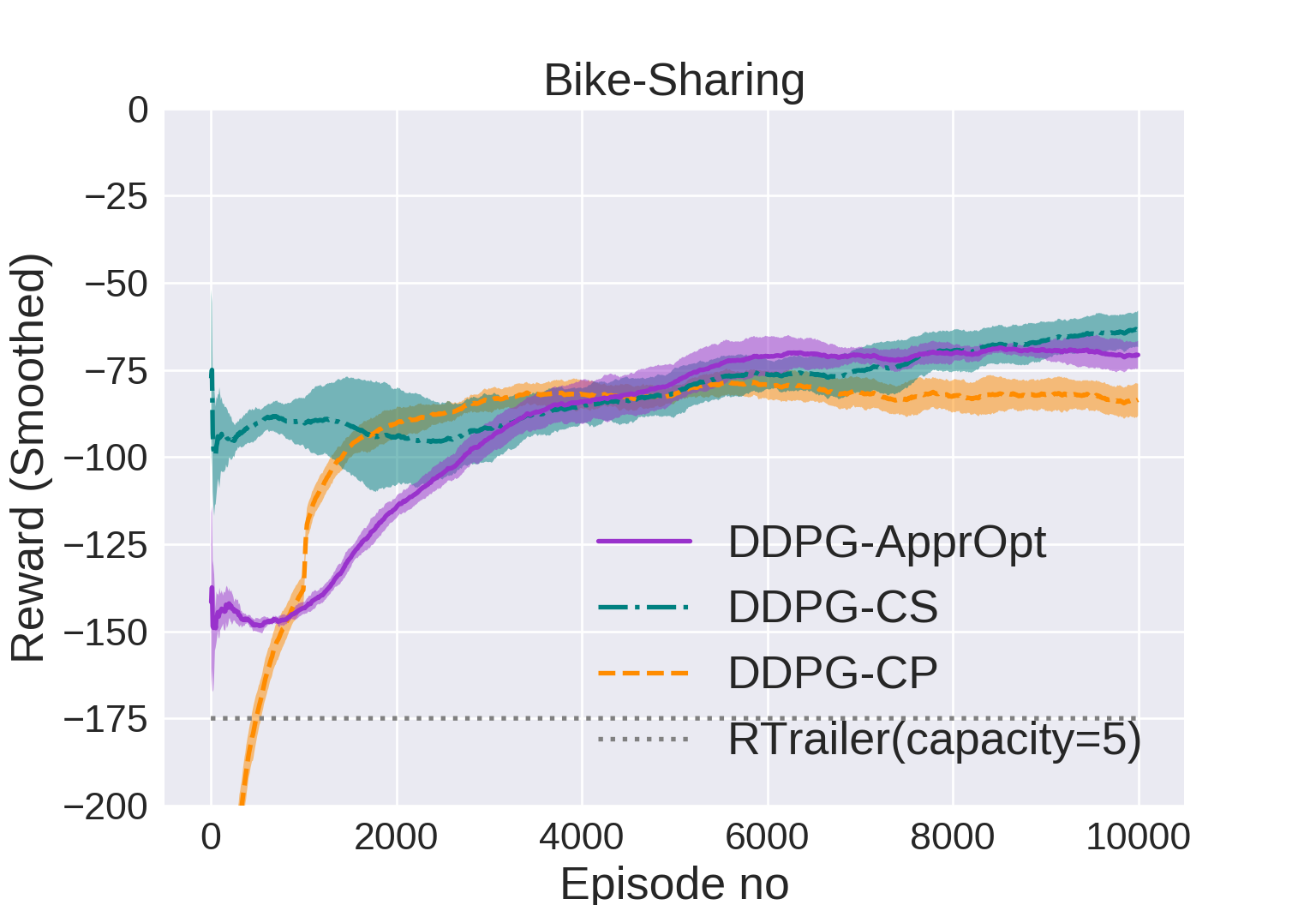}
    \includegraphics[scale=0.19]{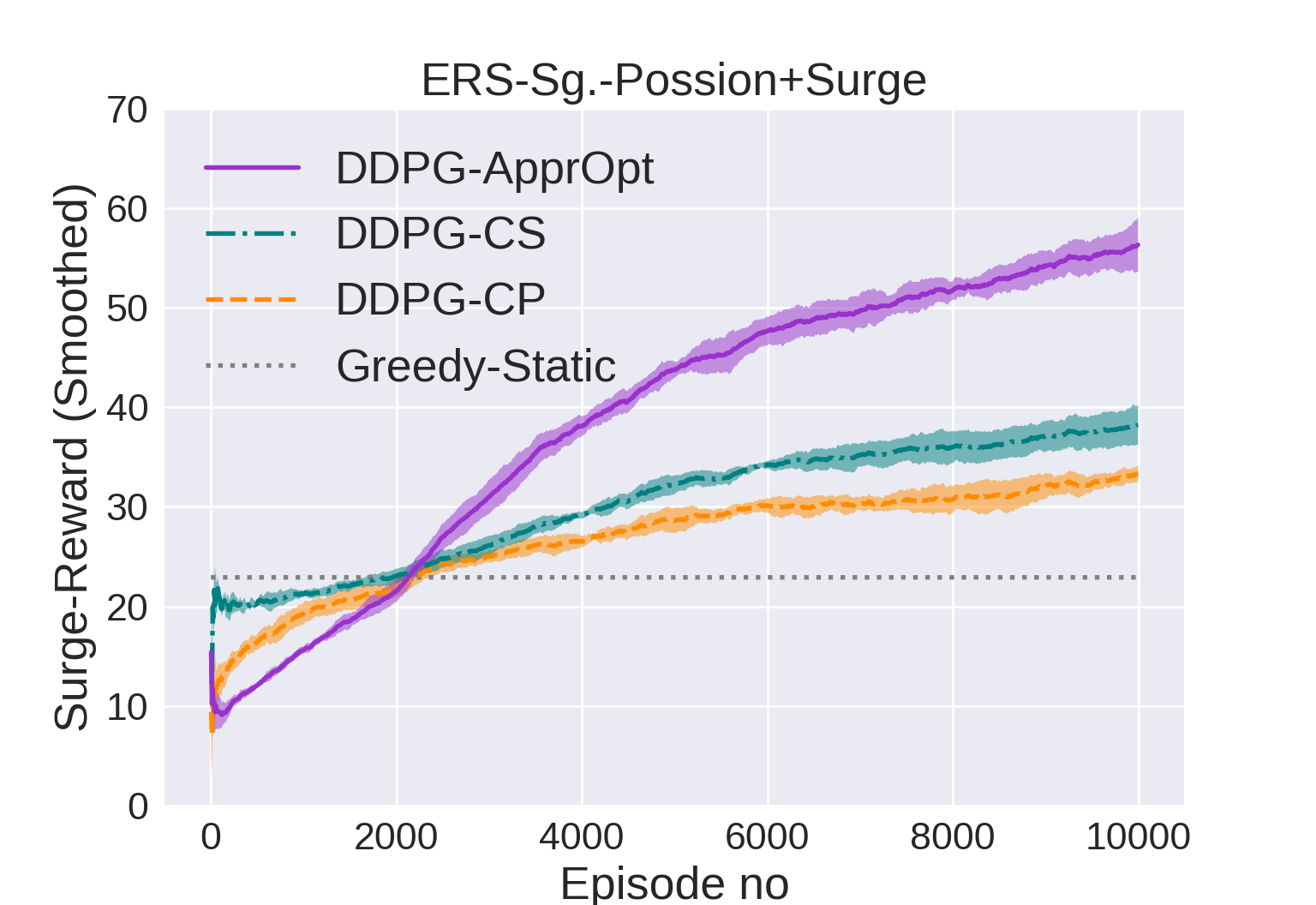}
    \centering
    \caption{\small Learning curves comparing different approaches. Each curve shows average reward per episode $\pm$ standard deviation over different seeds. The last subfigure shows the average reward \textit{due to the incidents which were part of the surges}.}
    \label{fig:learning-curves}
\end{figure*}

\begin{table}[t!]
        \begin{tabular}{l||r|r|r|r}
                Domain (Demand pattern) & Baseline & DDPG-CP & DDPG-CS & DDPG-ApprOpt\\
                \hline
                ERS Poisson & $342.9$ & $342.09\pm0.35$ & $339.42\pm1.35$ & $341.16\pm0.84$ \\
                ERS Poisson+Surge & $309.3$ & $316.46\pm1.38$ & $312.32\pm3.59$ & $353.20\pm4.17$\\
                BS & $-175$ & $-80.56\pm5.4$ & $-59.14\pm3.22$ & $-69.264\pm5.76$\\
        \end{tabular}
        \centering
	\caption{\small Average evaluation scores $\pm$ standard deviation over different seeds. The baseline approaches are Greedy-static for emergency response domain and RTrailer for BS (Bike Sharing) domain}
	\label{table:test-scores}
\end{table}

A summary of the results is presented in table \ref{table:test-scores}. Figures \ref{fig:learning-curves} show the learning curves for emergency response and bike sharing domains in different scenarios. 

The scores for the ERS domain instance with Poisson demand pattern reflect that all three of our approaches, DDPG-CS, DDPG-CP and DDPG-ApprOpt are competitive to the performance of greedy. This is in spite of greedy providing near optimal solutions in this domain as the environment is stationary. Additionally, greedy does not consider the bound constraints and hence the results are an upper bound on the actual solution that is achievable with constraints. 

All three approaches outperform the baseline in the ERS domain instance with random surges. As expected, the difference is mainly due to RL performing much better over the surges period of the episodes, which owing to their unpredictability, cannot be handled by an offline and static approach. DDPG-ApprOpt is particularly good in handling surges and is able to very significantly outperform all other approaches. DDPG-CS and DDPG-CP had roughly the same performance with DDPG-CP having the slight edge.

On the bike sharing domain with real data, there is considerable unpredictability in demand and so RL vastly outperforms the baseline RTrailer, which again being an offline approach cannot handle unpredictability so well. Also, it needs to be noted that all the RL approaches generalized very well to the test environment, which uses different data than the training environment. DDPG-CS performed the best and DDPG-ApprOpt was a close second. 

In summary, the results suggest that:
\squishlist
\item DDPG-ApprOpt has a clear edge over the other approaches performing either competitively or significantly better than other approaches on all the benchmark problems and over multiple seeds of training and testing. 
\item DDPG-CS came second, but it has the disadvantage of being limited to only cases where $\forall k, \epsilon_k > 0$. 
\item Finally, while DDPG-CP never wins clearly, it is very general, efficient and easy to implement, which might make it desirable in some scenarios.
\squishend

\section{Conclusion}
In summary, we have shown how RL may be used in online resource allocation problems where traditionally offline approaches or heuristics were used due to exponential action spaces and inability of RL to handle constraints. We presented three novel approaches based on DDPG for the same. We showed that specially in settings with non-Poisson demand patterns, RL has an important value due to its ability to have a reactive policy based on the situation. We backed up our claims with empirical evidence gathered by testing our approaches on simulators based on emergency response and bike sharing domains, using real or semi-real data. Each approach seemed to have its own pros and cons in terms of generality, efficiency and effectiveness.

\bibliography{references}
\bibliographystyle{aaai}

\end{document}